\documentclass[letterpaper, 10 pt, journal, twoside]{ieeetran}

\IEEEoverridecommandlockouts                              

\usepackage[english]{babel}

\usepackage{graphicx}
\usepackage{animate}
\usepackage{epsfig} 				
\usepackage{epstopdf}

\usepackage{listings}
\usepackage{color}
\usepackage{nameref}
\usepackage{hyperref}
\usepackage{amsmath}	 			
\usepackage{amssymb}  				

\usepackage{dsfont}			
\usepackage{mathtools}

\usepackage{epigraph}
\usepackage{lscape}
\usepackage[]{nomencl}				
\usepackage{algorithm}
\usepackage{algorithmic}
\usepackage{multicol}
\usepackage{multirow}
\usepackage{etoolbox}

\usepackage{caption}
\usepackage{subcaption}
\usepackage{wrapfig}

\usepackage{siunitx}

\usepackage{floatflt}

\usepackage{url}

\usepackage{enumitem}
\usepackage{placeins}

\newcommand{\email}[1]{\href{mailto:#1}{\nolinkurl{#1}}}

\newcommand{\link}[1]{\colora{\url{#1}}}

\renewcommand{\sec}[1]{Section~\ref{#1}}
\newcommand{\fig}[1]{Fig.~\ref{#1}}

\newcommand{\tab}[1]{Table~\ref{#1}}

\newcommand{\revision}[1]{#1}

\newcommand{\website}[0]{\url{https://github.com/facebookresearch/tacto}}
\newcommand{\simulator}[0]{TACTO}

\usepackage{bm}

\newcommand{\citet}[1]{\cite{#1}}
\newcommand{\citep}[1]{\cite{#1}}

\title{\simulator:\\ A Fast, Flexible, and Open-source Simulator for High-Resolution Vision-based Tactile Sensors}

\author{Shaoxiong Wang$^{1,\ast}$, Mike Lambeta$^{2}$, Po-Wei Chou$^{2}$, and Roberto Calandra$^{2}$%
\thanks{Manuscript received: September, 9, 2021; Revised December, 16, 2021; Accepted January, 9, 2022.}
\thanks{This paper was recommended for publication by Editor Dan Popa upon evaluation of the Associate Editor and Reviewers' comments.}%
\thanks{$^\ast$ Work done during an internship at Meta AI.}
\thanks{$^{1}$ Massachusetts Institute of Technology}%
\thanks{$^{2}$ Meta AI, Menlo Park, CA, USA}%
\thanks{Corresponding author: Roberto Calandra (\email{rcalandra@fb.com})}%
\thanks{Digital Object Identifier (DOI): 10.1109/LRA.2022.3146945}
}

\DeclareCaptionFont{8pt}{\fontsize{8pt}{8pt}\selectfont}
\begin{document}

\renewcommand{\figurename}{Fig.}

\maketitle


\begin{abstract}
	Simulators perform an important role in prototyping, debugging, and benchmarking new advances in robotics and learning for control.
Although many physics engines exist, some aspects of the real world are harder than others to simulate.
One of the aspects that have so far eluded accurate simulation is touch sensing.
To address this gap, we present \simulator{} -- a fast, flexible, and open-source simulator for vision-based tactile sensors.
This simulator allows to render realistic high-resolution touch readings at hundreds of frames per second, and can be easily configured to simulate different vision-based tactile sensors, including DIGIT and OmniTact.
In this paper, we detail the principles that drove the implementation of \simulator{} and how they are reflected in its architecture. 
We demonstrate \simulator{} on a perceptual task, by learning to predict grasp stability using touch from 1 million grasps, and on a marble manipulation control task. 
Moreover, we provide a proof-of-concept that \simulator{} can be successfully used for Sim2Real applications.
We believe that \simulator{} is a step towards the widespread adoption of touch sensing in robotic applications, and to enable machine learning practitioners interested in multi-modal learning and control.
\simulator{} is open-source at \website{}.
\end{abstract}



\section{Introduction}
	
	\IEEEPARstart{S}{imulators} play an important role in prototyping, debugging and benchmarking new advances in robotics.
With an appropriate simulator, expensive and time-consuming experiments in the real world can be approximated inexpensively on our computers. 
This allows to perform orders of magnitude more experiments at a fraction of the effort, and in many cases of the time.
The robotics community has traditionally made extensive use of simulators for control, and many different physics engines are available to researchers and practitioners~\citep{Erez2015Simulation}.
One aspect that has proven so far to be difficult to simulate is tactile sensing, and in particular vision-based tactile sensors~\citep{Yuan2017GelSight,Padmanabha2020OmniTact,Lambeta2020DIGIT} which provide rich high-resolution measurements.
This is because to accurately model this family of tactile sensors it is necessary not only to model the dynamics of the contact, but also to model the optical properties of the sensors and the corresponding illumination to obtain realistic perceptual outputs.
All of this while keeping the simulator flexible enough to implement various sensors with different form factors, and fast enough to be of practical use.

\begin{figure}[t]   
    \centering
    \includegraphics[width=\linewidth]{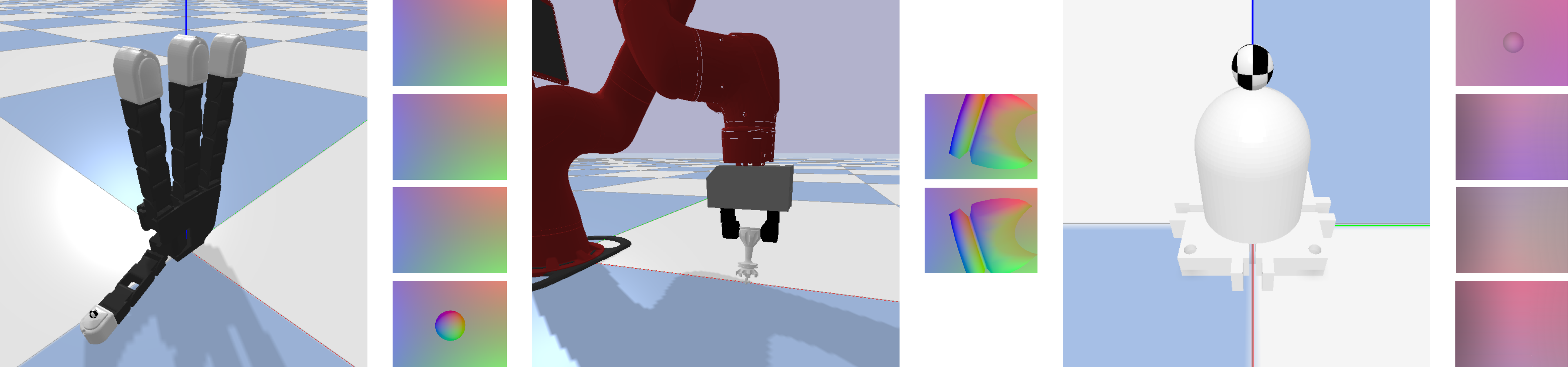}
    \caption{We open-source \simulator{} -- a simulator of vision-based tactile sensors. \simulator{} produces high-resolution and high-fidelity reading from tactile sensors at high-frequency (\SI{>100}{\hertz}). Its modular structure allows to model different vision-based tactile sensors and to be integrated with different physics engines. We believe that such a tool can benefit the touch sensing and robotic community, as well as researchers in machine learning that can now access a new sensor modality.}
    \label{fig:teaser}
\end{figure}

To fill this lack of touch sensing simulators, we introduce \simulator{} -- a simulator of vision-based tactile sensors explicitly designed to be fast and flexible. 
Our main contribution is to develop and open-source this simulator of vision-based tactile sensors.
In this paper, we describe the design choices adopted, the resulting software architecture of the simulator, and discuss some of its most important features.
Following, we present simulated experiments to demonstrate the capabilities of \simulator{} on perception and control tasks.
\fig{fig:teaser} shows examples of \simulator{} in different scenarios.

\simulator{} can natively be used in conjunction with PyBullet~\citep{CoumansPyBullet}, but can also be interfaced with other physics engines.
Particular care was dedicated to having a simulator that could be both fast and flexible.
Through our design, \simulator{} can render hundreds of frames per second, thus making the simulator practical for many control and learning to control applications.
In addition, thanks to its modularity it is easy to implement vision-based tactile sensors with different form factors and lighting properties.
Currently, \simulator{} implements two recent vision-based tactile sensors: OmniTact~\citep{Padmanabha2020OmniTact} and DIGIT~\citep{Lambeta2020DIGIT}.
While the ideal touch simulator would provide both realistic perceptual outputs and accurate contact dynamics, \simulator{} is aimed at tackling the rendering of realistic perceptual outputs and the creation of accurate contact dynamics is currently left to the underlying physical engine being used by the user.
We demonstrate our simulator by learning grasp stability models from touch, and by learning in-hand marble manipulation.
For learning grasp stability from touch, we collected a simulated dataset of 1 million grasps, which is orders of magnitude more data than the largest dataset previously used for this task~\citep{Calandra2017Feeling}.
This highlights the advantages of having a fast simulator of vision-based tactile sensors and allows us to better understand the performance of machine learning grasp stability models trained from large datasets that have, so far, been infeasible to collect in the real world.
Moreover, we apply Bayesian optimization to learn in simulation control to manipulate marbles between two fingers from touch. 

Finally, we present a proof-of-concept of Sim2Real on a  pose-estimation task to show the comparison between simulated and real signals, and various ways to improve the performance.
With the Sim2Real gap in mind, our major goal is to provide a playground, similar to OpenAI Gym~\citep{brockman2016openai}, to test different design or learning algorithms for the touch modality. The model can be later trained with real data. But some fundamental understanding and experience learned in simulation can be helpful for real robots, e.g. exploring the possible representation/policy to combine vision and touch for robotic tasks.

We believe that \simulator{} can be of practical value for different communities: 1) To hardware designers, it provides a preliminary method to simulate and evaluate their design choices for future sensors; 2) To the robotic community, it provides a way to simulate and study the integration of touch sensing into control scheme; 3) To the machine learning community, it provides an easy-to-use tool to generate multi-modal inputs which would otherwise require real-world hardware. 
To enable and stimulate researchers and practitioners in these fields, we open-source \simulator{} at \website{}.

	

\section{Related Work}
\label{sec:related}

	While there is a large number of physics engines available to robotic practitioners~\citep{Erez2015Simulation}, the choice when simulating tactile sensors is more limited.
This is mostly due to the difficulty of accurately and efficiently simulating touch.

Several traditional low-dimensional sensors have been simulated in the literature, including BioTac~\citep{Ruppel2018Simulation} and fabric-based tactile sensors~\citep{Melnik2019Tactile}.
\citet{Geukes2017Gazebo} simulated the iCub's iSkin in Gazebo through the use of an array of tactels appropriately distributed on the kinematic structure of the robot.
The speed of the simulation was however severely impacted by modeling such numbers of sensors independently.
Similarly, \citet{Habib2014SkinSim} made use of Gazebo to model general-purpose robotic skin, but modeled each element of the array of tactels as a spring-mass-damper system to provide an improved characterization of the mechanical properties of the skin.
In contrast to these works that simulate low-dimensional tactile sensors, we aim to simulate high-resolution vision-based tactile sensors, which can possess millions of tactels.
This creates novel challenges and requires different modeling approaches.

Exploiting the nature of vision-based tactile sensors, it is possible to use ray-tracing models from computer graphics to render sensor output. 
There are emergent simulators that shows promise for vision-based tactile sensors based on Phong's model~\citep{hoganseeing, Gomes2019GelSight}, Mitsuba2\cite{sodhi2020learning}, Unity~\citep{Ding2020Sim}, optical flow~\citep{Sferrazza2020Learning}, and finite elements models~\citep{kuppuswamy2019fast}. \revision{Among them, the closest concurrent works are \citep{sodhi2020learning, Gomes2019GelSight}. For comparison, the authors in \citep{sodhi2020learning} focused on simulating realistic tactile imprints using Mitsuba2 renderer\citep{nimier2019mitsuba}, but have not worked on efficiently integrating it with physical simulators so far. The authors in \citep{Gomes2019GelSight} combined the Phong’s model with the Gazebo simulator, but the experiments were mostly performed with a tactile sensor pressing on fixed objects. In contrast, we focus on exploring the scenarios where the sensors actively interact with the objects during the grasping or manipulation.} Overall, we aim to provide an open-source simulator that is fast, flexible, and can be efficiently integrated with the physics engine for experimenting with different learning and control algorithms with touch modality. 
	

\section{A Fast and Flexible Simulator of Vision-based Tactile Sensors} 
\label{sec:approach}

\begin{figure}[t]   
    \centering
    \includegraphics[width=\linewidth]{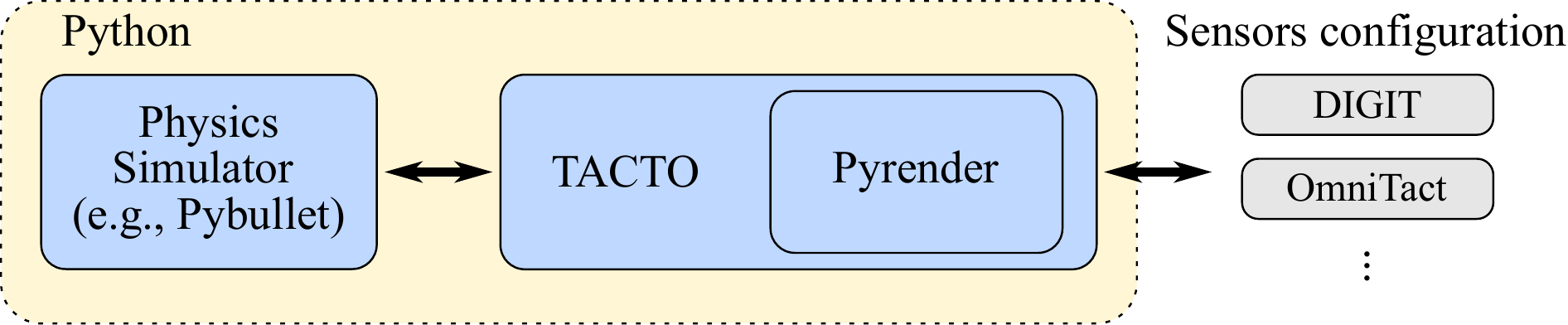}
    \caption{Software Architecture. \simulator{} bridges between physics simulator and back-end rendering engine, and can be configured to model different sensor designs through configuration files.}
    \label{fig:blockdiagram}
\end{figure}
\begin{figure}[t]   
    \centering
    \includegraphics[width=\linewidth]{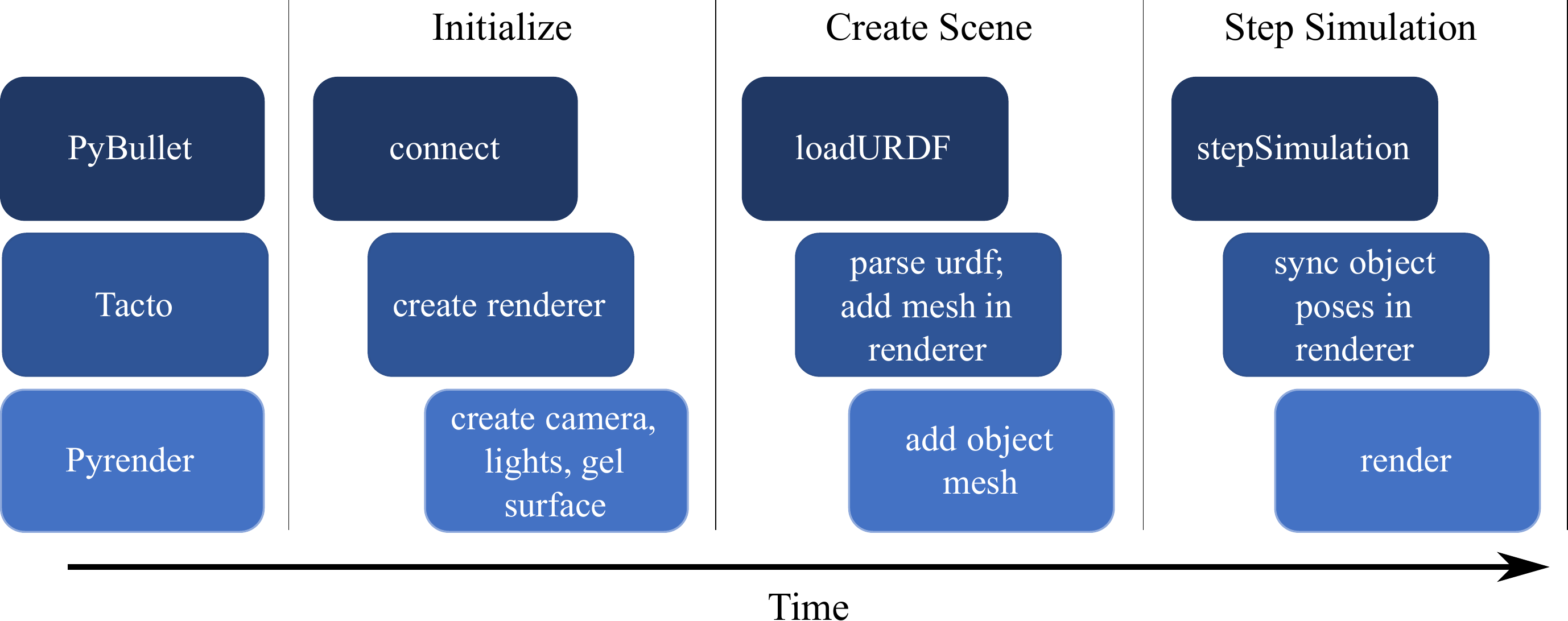}
    \caption{Workflow showing the functionality of \simulator{} at three major phases. (1) Initialize: create the sensor structure in the renderer; (2) Create scene: parse the objects URDF and add them into the  renderer; (3) Step simulation: synchronize the object poses from physics engine to the renderer.}
    \label{fig:workflow}
\end{figure}

We now detail the desired design considered when designing our simulators, and the corresponding architectural choices made to achieve these desiderata.
Following, we discuss some of the salient features that \simulator{} offers. 

\subsection{Design Desiderata}

\textbf{High-throughput:} simulators must be as fast as possible to reduce the real-world time of running simulations.
Reproducing touch sensing, even for low-dimensional sensors, has traditionally been a very computationally intensive operation often leading to simulations barely faster than real-time~\citep{Geukes2017Gazebo}.
Obtaining a simulator that could perform hundreds of frames per second was, from the beginning, one of our most important design desiderata. 

\textbf{Flexible:} since there are different sensor designs of vision-based tactile sensors, where some of them have complex geometry~\citep{Padmanabha2020OmniTact}, mirrors~\citep{donlon2018gelslim}, and transparent case for light piping~\citep{romero2020soft}, it is desirable for any simulator to be flexible and powerful enough to support a large choice of optical components, and mechanical designs.

\textbf{Realistic:} it is desirable that the simulator produces outputs as close as possible to real measurements, including illumination of the contact region, global lighting distribution, and details like shadows and deformation on the contact boundary.

\textbf{Easy to Use:} finally, it is from a practical point of view very important that the simulator is easy to install, use, modify and set up for different perception and control tasks.

\begin{figure}[t]   
    \centering
    \includegraphics[width=\linewidth]{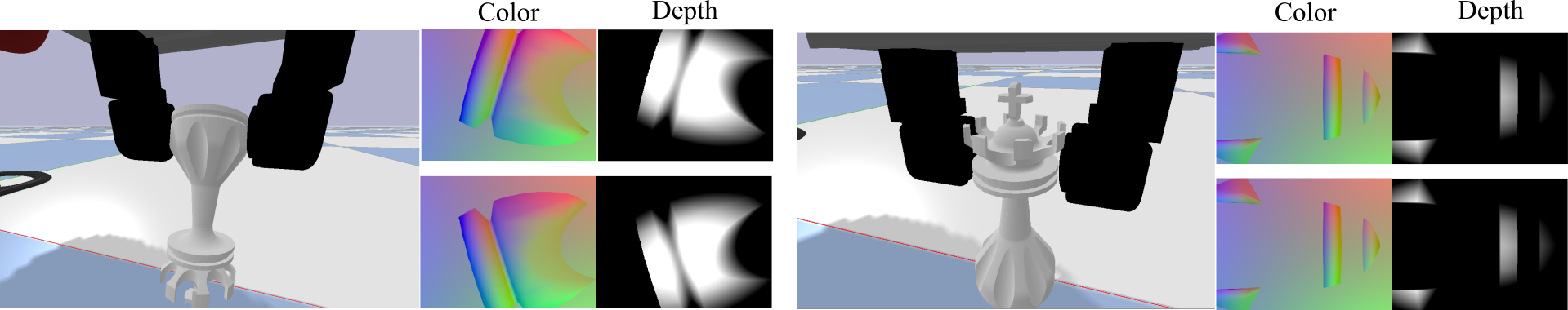}
    \caption{Example images of simulated DIGIT imprints. \simulator{} is able to generate color and depth images at the same time with details of the local geometry at high speed.}
    \label{fig:signal_digit}
\end{figure}
\begin{figure}[t]   
    \centering
    \includegraphics[width=\linewidth]{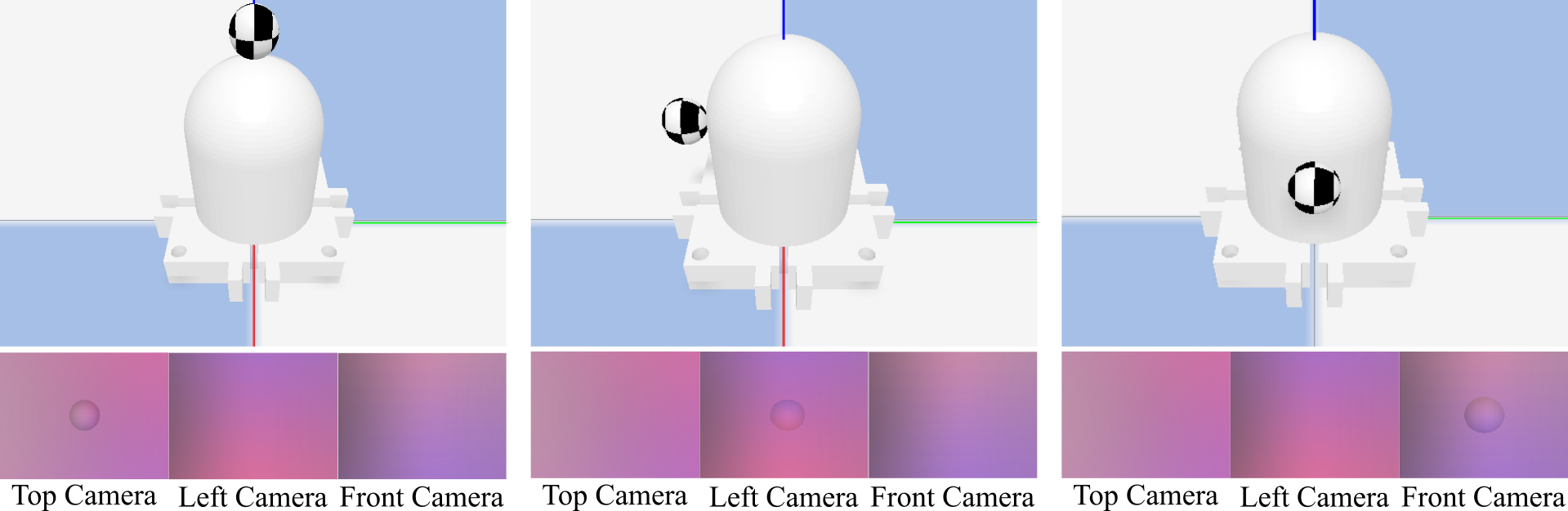}  
    \caption{Example images of a simulated OmniTact~\citep{Padmanabha2020OmniTact} touching a sphere. We show only 3 of the 5 cameras mounted on the sensor. It is visible how the specific light/camera placement in the OmniTact results in more pinkish images also in our simulator.}
    \label{fig:signal_omnitact}
\end{figure}

\begin{table*}[tb]
\centering
\begin{tabular}{|c|c|c|c|c|c|c|c|c|c|c|c|c|c|}
\hline
                     & \multirow{2}{*}{Res} & \multicolumn{4}{c|}{1 obj (1 in contact)} & \multicolumn{4}{c|}{100 objs (1 in contact)} & \multicolumn{4}{c|}{100 objs (10 in contact)} \\
                     &                             & Step        & Sync        & Render        & \textbf{FPS}        & Step         & Sync         & Render         & \textbf{FPS}        & Step          & Sync         & Render         & \textbf{FPS}        \\ \hline
\multirow{3}{*}{GPU} & 160$\times$120                     & 0.2         & 2.2         & 2.5           & \textbf{220}         & 5.8          & 2.4          & 2.8            & \textbf{200}           & 11.6          & 6.3          & 6.9            & \textbf{80}            \\ \cline{2-14} 
                     & 320$\times$240                     & 0.2         & 2.2         & 5.0           & \textbf{140}         & 5.8          & 2.5          & 7.6           & \textbf{100}            & 11.5          & 6.3          & 9.5            & \textbf{60}            \\ \cline{2-14} 
                     & 640$\times$480                     & 0.2         & 2.3         & 9.4           & \textbf{90}          & 6.1          & 2.6          & 8.6            & \textbf{90}            & 11.7          & 6.3          & 15.0           & \textbf{50}            \\ \hline
\multirow{3}{*}{CPU} & 160$\times$120                     & 0.4         & 4.2         & 11.8          & \textbf{60}          & 9.8          & 4.5          & 11.3           & \textbf{60}            & 19.5          & 10.7         & 16.7           & \textbf{40}            \\ \cline{2-14} 
                     & 320$\times$240                     & 0.4         & 4.2         & 25.8          & \textbf{30}          & 9.6          & 4.6          & 26.1           & \textbf{30}            & 18.0          & 9.8          & 26.2           & \textbf{30}            \\ \cline{2-14} 
                     & 640$\times$480                     & 0.4         & 4.5         & 78.3          & \textbf{10}          & 9.0          & 4.8          & 78.5           & \textbf{10}            & 17.2          & 10.4         & 80.1           & \textbf{10}            \\ \hline
\end{tabular}
\caption{Breakdown time of each stage (in millisecond) and overall speed (in frames per second, excluding physics simulation) for \simulator{} with PyBullet when simulating a DIGIT sensor with multiple objects (each mesh with 12K faces) in the scene. Step: PyBullet simulates physics; Sync: TACTO synchronizes the scene; Render: Pyrender renders the scene. Note that physics simulation can run asynchronously to speed up. CPU machine: Intel Core i7-6820HQ. GPU machine: Nvidia RTX 2080 Super GPU with Intel Core i9-9900K CPU.}
\label{tab:breakdown}
\end{table*}

\subsection{Architectural Choices} 
\label{sec:arch_choice}

Here we compare several architectural choices. The simulator needs to calculate the local contact geometry (Depth), and the corresponding rendering (RGB) that best meets the desiderata. 


\textbf{1. Phong's model for RGB rendering from Depth (simple but less powerful):} 
PyBullet built-in camera can provide a depth map of the contact area. To render the RGB image from the depth map, researchers from \citep{Gomes2019GelSight} implemented their own renderer based on Phong’s reflection model. 
It generated promising results and should be easy to use. 
However, it assumed that light only bounces once, directly from the gel surface to the camera.
Hence, it is difficult to adapt to existing and future sensor designs that require advanced functionalities, like reflection, refraction, and shadows with fast speed. 
Although it can be extended to support more ray-tracing functionalities, this may require non-trivial engineering time to re-implement methods with GPU acceleration which are already provided and tested by open-source graphics libraries, like OpenGL\citep{shreiner2013opengl}.

\textbf{2. OpenGL for RGB rendering from Depth (powerful but slow):}
Alternatively, we can leverage the power of OpenGL~\citep{shreiner2013opengl}.
Besides the features in \citep{Gomes2019GelSight}, OpenGL also supports mirrors, transparent objects, shadow, GPU acceleration, etc, which opens up the possibility for rendering advanced sensor design at high speed, with minimal effort.
To use the power of OpenGL, one can get the depth image from PyBullet first, and pass the depth map to OpenGL for rendering. 
The rendering alone is fast and can be sped up in GPU, however, I/O speed becomes the bottleneck. 
In preliminary experiments, a significant amount of time was spent on loading the mesh generated by the depth map into OpenGL, and this limited the overall speed of this method to only 20 frames per second, even on GPU. 

\revision{\textbf{3. OpenGL for RGB rendering from synchronized scenes (proposed, powerful and fast):}}
Our proposed system design builds a synchronized scene from the physical simulator, and directly renders both depth and RGB images in OpenGL. It can achieve high speed with powerful rendering functionalities. 

Specifically, to avoid the I/O bottleneck of loading the mesh from depthmaps repeatedly, \simulator{} preloads the gel surface and object meshes into the OpenGL scene, then keeps synchronizing their poses from the physical simulator. 
At each step, \simulator{} overlaps the original gel geometry with the contact object, with the respective poses, in OpenGL to extract depth and RGB images.
Although loading meshes is slow, it is very fast to change their poses and re-render afterwards.
In this way, the system can render at very high speed (up to 200 frames per second in our experiments).

Note that this is designed to speed up the computation for interacting the sensor with rigid objects, where the deformation of the object itself is negligible. 
Since the silicone gel is much softer than many everyday objects, the method can still be applied to these objects.
To manipulate very deformable objects, \simulator{} provided the functionality to render RGB from depth, as described in the second option, with slower speed though.

One limitation of the method is that it is difficult to model the deformation of the gel on the contact boundary, because the RGB and depth images are calculated at the same time. 
However, this mostly affects sharp edges, and can be approximated by smoothing objects' meshes during pre-processing. 
We can also add data augmentation~\citep{perez2017effectiveness} or refine the rendering afterward with generative models to make it more realistic~\citep{shrivastava2017learning, hoffman2018cycada}. 
Besides, we expect that contact boundary contains relatively little useful information compared to other aspects like contact location, contact mask, object contact pose, normal forces, etc.
It can still be a good playground to study different algorithms in simulation for touch.
Overall, we think it is worthwhile to trade this for speed.

For implementation, we use Pyrender~\citep{Pyrender} as the renderer in \simulator{} since it provides a lightweight python interface for deploying OpenGL with GPU support, making \simulator{} not only powerful, but also easy to use.

Based on extensive experiments and analysis of each method, we decided to use this third option. 
This results in \simulator{} being closely aligned to our ideal simulator, which is fast, flexible, and powerful.
In \sec{sec:result}, we will demonstrate that \simulator{} is also easy to set up for perception and control tasks.

\subsection{Overview of the Software Architecture}

\fig{fig:blockdiagram} shows the overall software architecture of \simulator{}.
\simulator{} takes the responsibility of bridging between the physics simulator and the back-end rendering engine. 
Note that the architecture we propose leverages the advanced ray-tracing tools, which enables simulating high-quality tactile signals efficiently, with minimal efforts.
It can render different sensors by loading corresponding sensor configurations, which include parameters such as the camera, lights, and gel surface. 

\fig{fig:workflow} shows a summary of the overall workflow. \simulator{} includes three major phases (for simplicity, we here assume PyBullet as the underlying physics engine). 
1) Initialize:  \simulator{} loads the sensor configuration, and setup up the sensor (camera, lights, gel mesh) in the rendering engine.
2) Create scene: PyBullet loads object URDF into the scene, and \simulator{} parses the URDF (by urdfpy package in our case), and adds the analyzed mesh into the rendering engine. 
3) Step simulation: PyBullet first calculates physics simulation. Then \simulator{} loads the poses of each link from PyBullet, synchronizes the poses of objects and sensors in the rendering engine, and fetches the rendered tactile imprints.

\subsection{Salient Features}
\label{sec:features}

\textbf{Fast:}
\simulator{} is very fast for being a tactile sensor simulator.
\tab{tab:breakdown} shows the speed of \simulator{}. 
When interacting with an object mesh with 12K faces, \simulator{} is able to render a single DIGIT sensor at 200 frames per second on GPU with an output resolution of 160x120 pixels. 
It can also render multiple sensor outputs, e.g., rendering four DIGIT sensors mounted on an Allegro hand at 50 frames per second on GPU with an output resolution of 160x120 pixels. 
We optimized the \simulator{} so that only the number of objects in contact influences Pyrender's speed, making it maintain a high speed even in a cluttered environment.
The rendering time grows linearly with higher output resolution on GPU, and the speed stays the same when the mesh size scales from 2K to 12K faces.

\begin{figure}[t]   
    \centering
    \includegraphics[width=\linewidth]{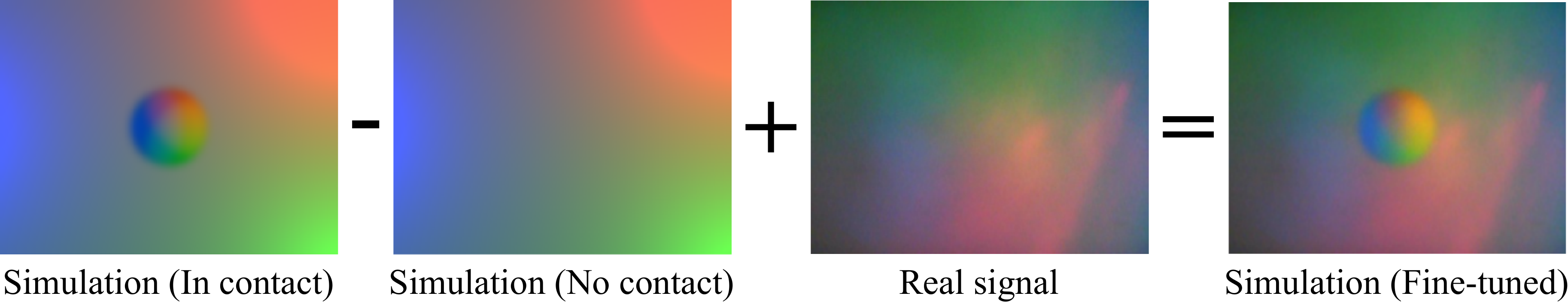}
    \caption{If readings from a real-world sensor are available, \simulator{} allows to fine-tune the simulator using the real-world data. This is achieved by calculating the pixel-wise difference of the simulated images with and without touch, and then adding the reference real-world image.}
    \label{fig:finetune}
    \end{figure}
    
\begin{figure}[t]  
        \centering
    \includegraphics[width=\linewidth]{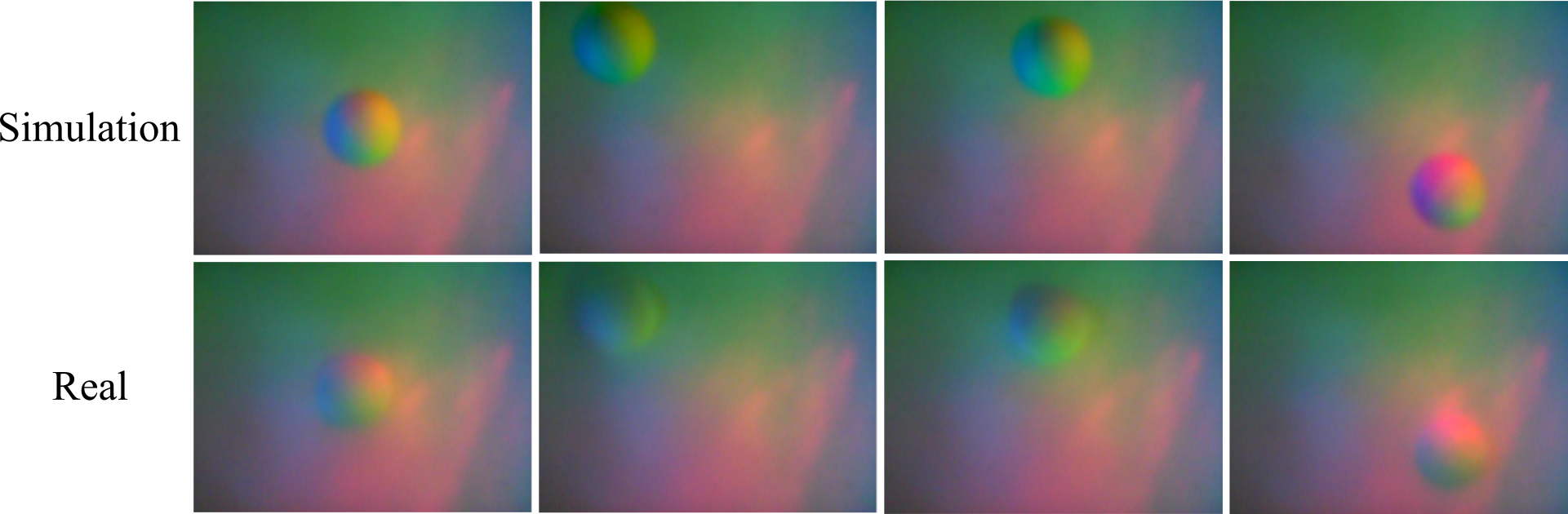}  
    \caption{Comparison of simulation and real signals with contacts across the sensor. \simulator{} captures the non-uniform light distribution similar to the real signals. The real-world readings are collected from a DIGIT sensor~\citep{Lambeta2020DIGIT} touching a ball of \SI{5.3}{\milli\meter} diameter. Simulated images are fine-tuned with a background real-sensor image.}
    \label{fig:simreal}
        \end{figure}
        
\begin{figure}[t]  
        \centering
    \includegraphics[width=\linewidth]{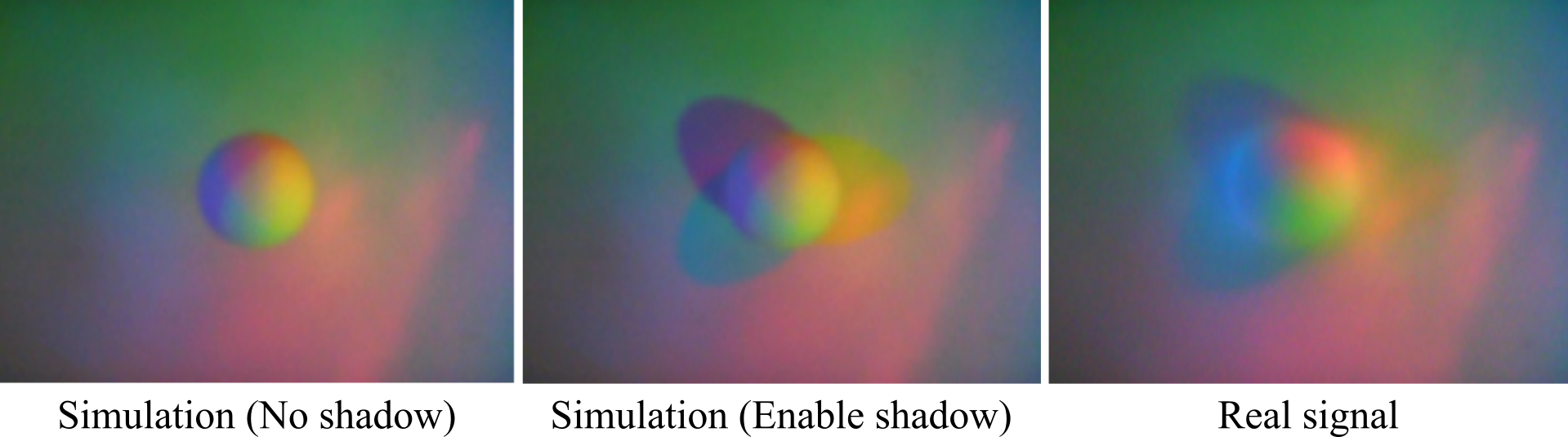}
    \caption{\simulator{} supports rendering shadows to obtain more realistic simulations. The real-world measurement is collected from a DIGIT sensor touching a ball of \SI{3.7}{\milli\meter} diameter.}
    \label{fig:shadow}
\end{figure}

\begin{figure*}[t]
    \centering
    \includegraphics[width=\linewidth]{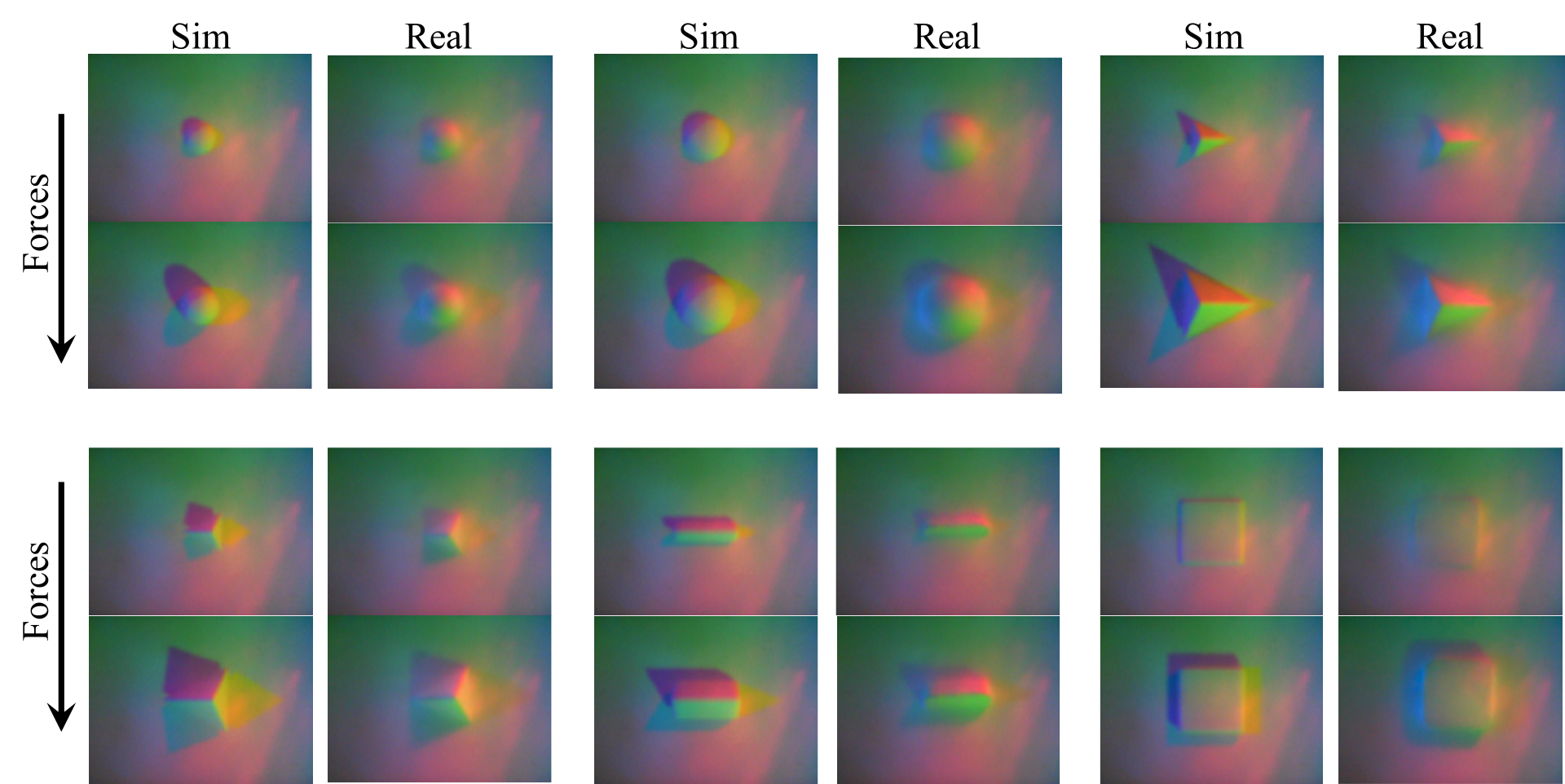}
    \caption{Comparison of simulated and real tactile readings.  Real readings are collected from a DIGIT sensor touching objects with different shapes (spheres, corners, edges, planes), with different forces. \simulator{} generates realistic tactile signals for both contact surfaces and shadows. The simulated images are overlaid on a background real sensor image. We smooth the cube's mesh to better model the deformations on the contact boundaries, during pre-processing. A 15$\times$15 Gaussian filter is applied on the rendered RGB image to match the imperfection from camera/light source in the real sensor. Note that it takes only 2 extra milliseconds for rendering shadows (with meshes of 19k faces, output resolution of 640x480, GPU).}
    \label{fig:comparison}
\end{figure*}

\textbf{Flexible:}
\simulator{} can adapt to different sensor designs by changing configuration files easily. Besides rendering DIGIT~\citep{Lambeta2020DIGIT} signals, as shown in \fig{fig:signal_digit}, we also demonstrate the possibility to render OmniTact~\citep{Padmanabha2020OmniTact} signals, as shown in \fig{fig:signal_omnitact}, which has a substantially more complicated sensor structure including round surface, 5 cameras, and 11 light sources. 
For the configuration file details, it requires parameters of the gel (pose, mesh), a list of camera(s) (pose, field of view, clipping plane), a list of lights (pose, color, intensity), and user-specific details like noise level, and the mapping from force to deformation. 
The example configuration files in the repository include the explanations for each parameter. 
The configuration file and renderer can be further extended to support more functionality of OpenGL/Pyrender and sensors.

\textbf{Force dependent:}
\revision{In \simulator{}, the contact forces are generated from the physics engine of choice (currently by default PyBullet with a rigid body contact model).}
\revision{However, within \simulator{} we also apply a deformation function to simulate the dynamics range of the deformation of the gel by mapping the force measured in the physics engine to different deformations of the mesh used to simulate the gel.}
This means that light forces will yield fewer deformations, and higher forces to more deformation of the gel.
The current deformation function used is a piece-wise linear mapping from normal forces to deformation depth, which approximates the linear elasticity of real sensors within a reasonable range~\cite{ma2019dense} and consider both a lower (below which the sensor is not capable of sensing deformation) and an upper threshold (above which the deformation is saturated) to the force sensed. However, it is straightforward for the end-user to provide their own deformation function, for example, by characterizing the gel with a non-linear function. In the future, we plan to provide more realistic transfer functions for the sensors to which we have access.

\textbf{Rendering from depth:}
\simulator{} also supports rendering tactile imprints from depth images as an option. Given the depth image, it generates corresponding meshes to replace the original gel surface in the rendering engine. 
As mentioned in \sec{sec:arch_choice}, the speed is limited due to I/O bottleneck, however, it can be helpful in some cases where it requires modification on depth images before rendering for more realistic tactile imprints.

\textbf{Calibration from real sensors:}
To make the rendering more realistic the simulator also supports a procedure, sketched in \fig{fig:finetune}, that allows to fine-tune the rendering using readings collected from real-world sensors. 
This results in highly realistic renderings that can be easily customized to each sensor.
In addition, \simulator{} captures the illumination changes across the sensor similar to real measurements. 
In real sensors, the light becomes dimmer when traveling longer, which generates non-uniform light distributions.
\fig{fig:simreal} shows a comparison across different contact regions. 
Finally, \simulator{} supports rendering shadows to match real signals, as shown in \fig{fig:shadow}. 
This option is easy to enable with our framework, and takes only $\approx2$ extra milliseconds to render on GPU. 
In contrast, it can take non-trivial time to re-implement shadow rendering from scratch with Phong's reflection model~\citep{Gomes2019GelSight}, especially to optimize for GPU acceleration.
\fig{fig:comparison} shows a comparison between simulated and real imprints. 
\simulator{} can simulate realistic rendering for both contact surfaces and shadows, with different contact geometry and forces.

\textbf{Compatibility to different physics engines:} 
We use an open-source physics engine, PyBullet, to demonstrate the framework. 
But \simulator{} can also support other physics engines. 
Specifically, there are two ways: 1) is to synchronize the scene with the functions provided by the selected physics engine. 
The required functions include getting object/link poses to synchronize the scene, and getting contact forces to simulate deformation. 
2) is to render from depth as described in \sec{sec:features}. 
Provided the depth information, \simulator{} can generate the mesh in the scene and render corresponding images.


\section{Simulated Experiments}
\label{sec:result}

\begin{figure*}[t]   
    \centering
    \includegraphics[height=5cm]{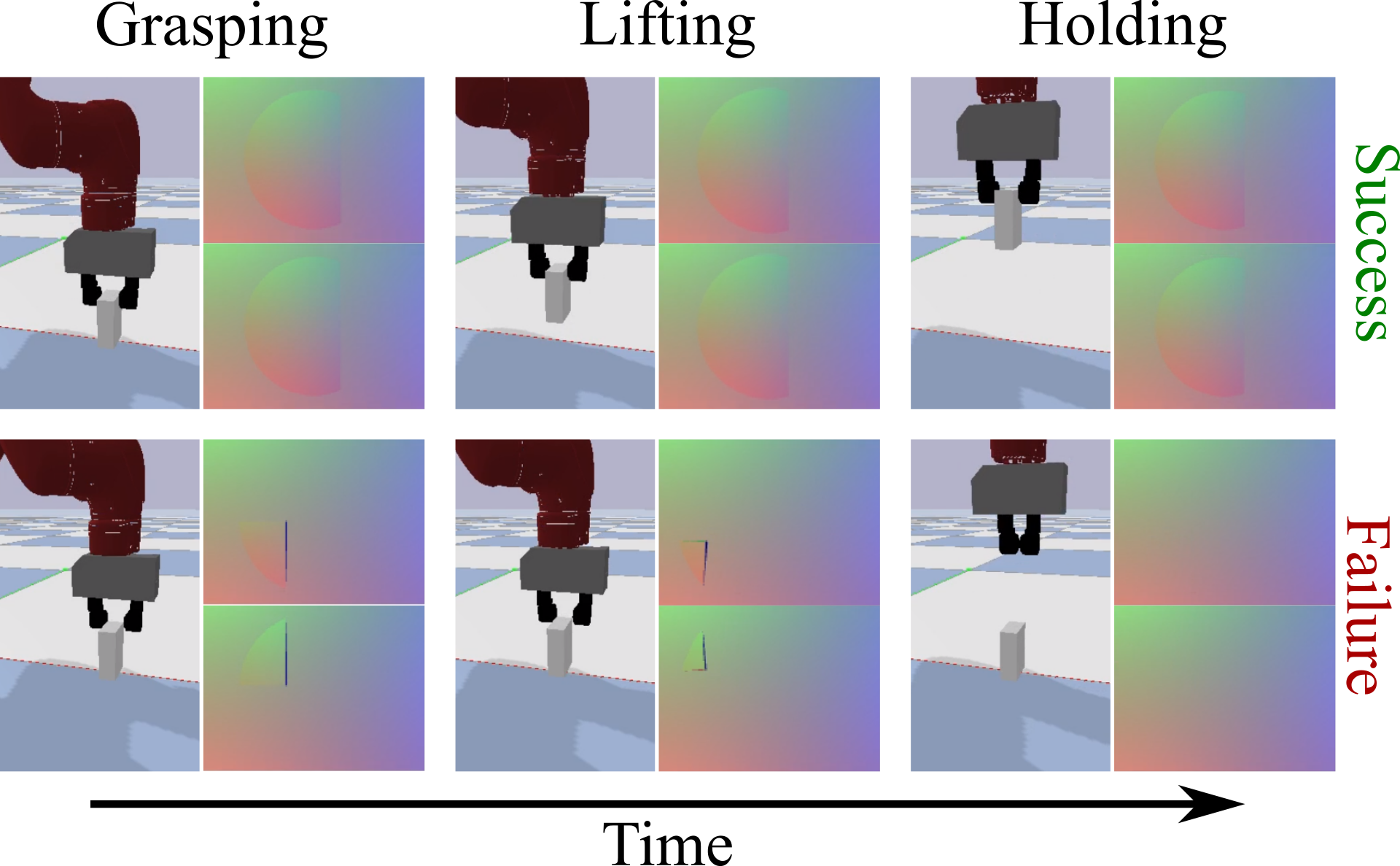}
    \hspace{10pt}
    \includegraphics[height=5cm]{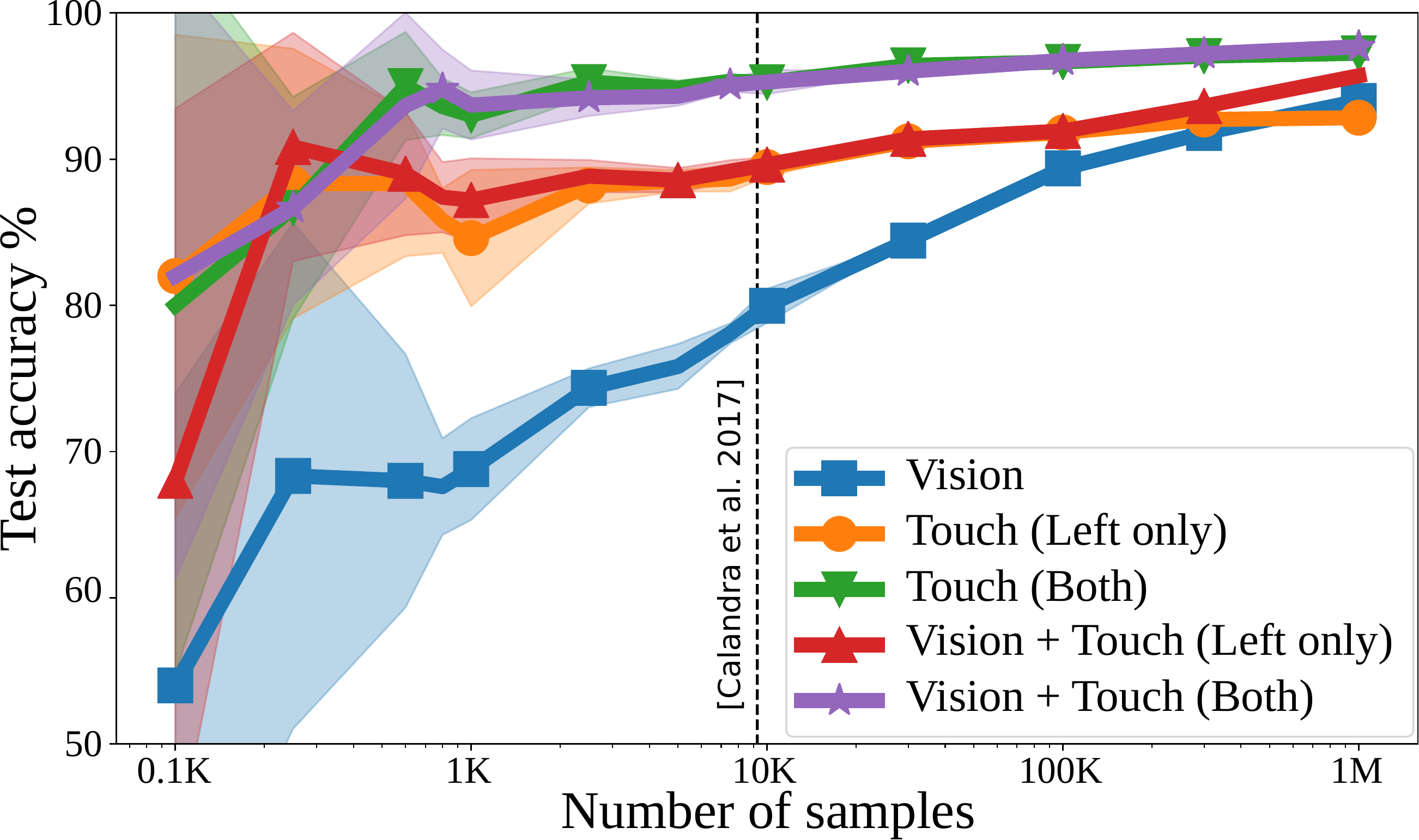}
    \caption{Learning Grasp Stability. (\textit{Left}) Examples of a successful grasp and a failure grasp. In the failure grasp, the object is only grasped by the corner and begins to slip after being lifted. (\textit{Right}) Median and 68\% percentile of the learned models when varying the number of data used. We compare using only vision, only touch and both vision and touch as inputs of the models. Results show that learning grasp stability from touch needs significantly less amount of data to achieve relative high accuracy compared to vision, and that increasing the amount of data helps to improve performance. The vertical dashed line shows the largest dataset collected on real robot~\citep{Calandra2017Feeling}. In the simulation, we can experiment with a dataset more than two orders of magnitude larger.}
    \label{fig:stability}
\end{figure*}

We now demonstrate \simulator{} on a perception task to learn grasp stability from touch, and on a control task to manipulate a marble between two fingers using touch. We choose these two tasks based on previous works with real robots\cite{Calandra2017Feeling,tian2019manipulation} for better comparison between simulated and real environments. The experiments in simulation achieve similar results to the ones with real robots, which demonstrates the effectiveness and potentials of the simulated environment.
For these experiments, we use \simulator{} in conjunction with the PyBullet physics engine~\citep{CoumansPyBullet}.

\subsection{Learning Grasp Stability in Simulation}
In this task, we learn in simulation a classifier of grasp stability from vision and touch readings, following the previous work on real robots~\citep{Calandra2017Feeling}. The goal is to predict whether a grasped object will be successfully lifted, based on the touch and vision readings before the lifting. We aim to investigate whether the results in simulation match the real ones in \citep{Calandra2017Feeling}.

\begin{figure*}[t]   
    \centering
    \includegraphics[width=0.82\linewidth]{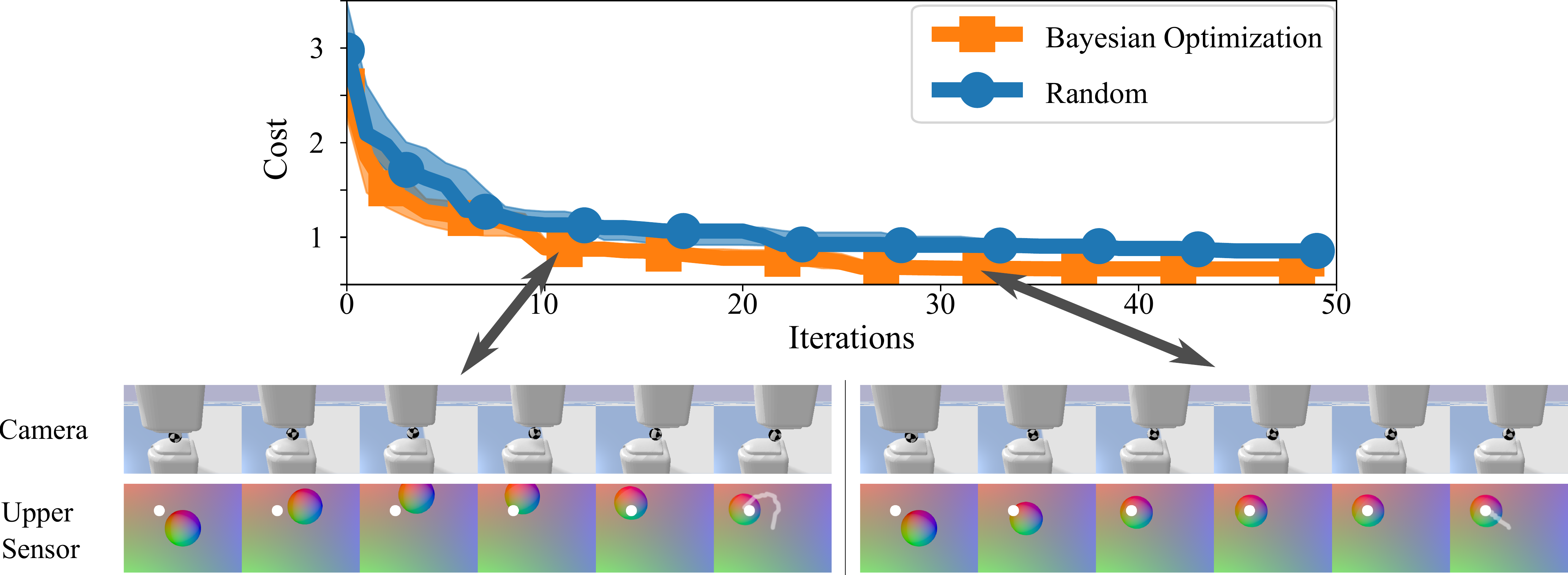}
    \caption{Learning in-hand marble manipulation. (\textit{Upper}) The learning curve (median and 68th percentile over 30 experiments) for Bayesian optimization and random search. (\textit{Lower}) Examples of a marble rolling into different target locations within the fingers at the early and later learning stage. At the later stages of learning, the controller can roll the marble between fingers with faster speed and higher accuracy, while avoid dropping the marble. }
    \label{fig:marble}
\end{figure*}

\textbf{Setting}
Our setup consists of two DIGIT sensors~\citep{Lambeta2020DIGIT} mounted on a WSG-50 parallel-jaw gripper, and an external camera.
We collected 1 million grasps in a self-supervised manner by randomizing the position, orientation and force applied by the gripper. 
The ground truth for each grasp was labeled depending on whether the object was still between the fingers after being lifted. Some examples are shown in \fig{fig:stability}.
It took only one day to collect 1 million grasps with 5 threads. 
In our experiment, we used a single box object for demonstration purposes, but it would be straightforward to extend to various object datasets, such as YCB dataset~\citep{calli2015ycb}, or Dex-Net~\citep{mahler2017dex} dataset.
The resulting dataset collected with \simulator{} is several orders of magnitude larger than any publicly available dataset of grasps using tactile sensing~\citep{Calandra2017Feeling}.
As such, it allows us to evaluate the performance of learned grasp stability models in data regimes that are still unexplored.
To learn the grasp stability, we trained ResNet-18~\citep{he2016deep} neural network models that, given the raw vision and touch signals, would predict whether the grasp is stable or not after lift-off. 
The training procedure followed previous work~\citep{Calandra2017Feeling}. 
For details, to fuse vision and touch signals, the feature vectors produced by ResNet-18 were concatenated and fed to two fully connected layers, with 512 and 256 hidden units, to predict final results. 
To speed up training, we used ResNet-18 model pre-trained on ImageNet~\citep{deng2009imagenet}. 
And we used vision and touch images with $160\times120$ pixels. The images were resized to $256\times256$ and randomly cropped to $224\times224$ for data augmentation.

To evaluate the performance of different dataset sizes, we used K-fold cross-validation and computed the median and 68\% percentile of the classification accuracy. 
Due to computational limits, we used $K = 10$ up to 1000 datapoints, $k=5$ up to 10k datapoints, and above 10k datapoints we evaluated a single train/test split $80/20 \%$.
We trained 10 epochs for each dataset size using Adam optimizer~\citep{kingma2014adam} with a learning rate of $5e^{-4}$ and batch size of $32$. 

\textbf{Results}
The results shown in \fig{fig:stability} suggests:
(1) the model learned fast from touch with only a little amount of data, while vision requires 3 or 4 orders of magnitude more data to catch up; 
(2) single tactile sensor worked significantly worse than two tactile sensors, because the object may look stable from one tactile sensor while it unstably contacts the other side;
(3) on the low-data regime, the result agrees with previous real-world experiments~\citep{Calandra2017Feeling}: combining vision and touch worked best in most of the cases. 
Although touch-only achieved comparable results to the combined model, we think the difference can increase with a larger object set; 
on the high-data regime, we can evaluate with 2 orders of magnitude more data in simulation compared to \citep{Calandra2017Feeling}, and observe the trends that all the models keep improving still, and vision's potentials with more data.

\subsection{Learning In-hand Marble Manipulation}

\revision{In this task, we learn in simulation to roll a marble to target locations in the sensor coordinate, following the previous work on real robots~\citep{tian2019manipulation}.
The setup includes two DIGIT sensors as shown in \fig{fig:marble}. 
The lower sensor is fixed, while the upper sensor is controlled to roll the marble. Since there are rich contacts with friction happening during the rolling, we aim to investigate how stable the \simulator{} and PyBullet are, and explore whether we could achieve similar performance to the real one in \citep{tian2019manipulation}.}

\textbf{Setting}
We use position control with a maximal force for controlling the upper sensor. Specifically, we control the horizontal position of the upper sensor, and push the sensor vertically with maximal force to keep the marble in hand while rolling.
We parameterize the controller as $\mathbf{u}=\mathbf{K}\mathbf{\bar{x}}$, where $\mathbf{u} \in {\mathbb{R}}^2$ is the desired velocity of the upper sensor in horizontal plane, $\mathbf{\bar{x}}\in{\mathbb{R}}^2$ is the error state between the current marble center location $\mathbf{x}$ and the goal location $\mathbf{x^*}$ in tactile space, and $\mathbf{K}\in{\mathbb{R}}^{2\times2}$ is the parameter to learn. The cost is defined as cumulative error distance in tactile space $\sum_t{\lVert \mathbf{\bar{x}_t} \rVert}$, and we set eight different target locations and take the average cost for robustness. We apply Bayesian optimization~\citep{snoek2012practical} with upper confidence bound to optimize the parameter $\mathbf{K}$ automatically.
Our main purpose here is to validate the simulation system, and provide benchmark experiments, however, the controller can be replaced by model predictive control and/or reinforcement learning to manipulate more complex objects~\citep{tian2019manipulation, akkaya2019solving} and for dexterous hands~\cite{Lambeta2020DIGIT}.

\textbf{Results}
\fig{fig:marble} shows the quantitative and qualitative results of rolling a marble into desired locations. The system can learn to roll the marble into different target locations with few iterations and roll faster with more iterations. During the experiments, we validate that both PyBullet and \simulator{} run as expected without abnormal situations. Because the simulation is fast, it only takes 8 minutes for Bayesian optimization to learn marble manipulation with 50 iterations. 
It includes 6 minutes for optimizing the acquisition function, and 2 minutes for simulation, where there are 50 iterations, and each iteration includes 50 steps for rolling into each of 8 directions, rendering 20,000 tactile imprints of $160\times120$ resolution in total.


\section{Sim2Real Experiments}
\label{sec:discussion}

\begin{table}[tb]
\centering
\begin{tabular}{l|c|c}
                                & \textbf{\begin{tabular}[c]{@{}c@{}}position error\\ (mm)\end{tabular}} & \textbf{\begin{tabular}[c]{@{}c@{}}rotation error\\ (degrees)\end{tabular}} \\ \hline
Random                          & \multicolumn{1}{l}{11.75 $\pm$ 1.15}                                   & \multicolumn{1}{|l}{46.56 $\pm$ 5.91}                                        \\
Sim2Sim                         & 0.41 $\pm$ 0.01                                                        & 3.48 $\pm$ 0.34                                                             \\ \hline
Real2Real (16 datapoints)                  & 4.45 $\pm$ 0.86                                                        & 33.85 $\pm$ 1.07                                                            \\
Real2Real (32 datapoints)                  & 3.48 $\pm$ 0.56                                                        & 25.45 $\pm$ 2.07                                                            \\
Real2Real (64 datapoints)                  & 2.01 $\pm$ 0.14                                                        & 10.96 $\pm$ 0.24                                                            \\
Real2Real (128 datapoints)                 & 0.76 $\pm$ 0.07                                                        & 4.96 $\pm$ 0.70                                                             \\ \hline
Sim2Real (without augmentation) & 4.56 $\pm$ 0.40                                                        & 17.64 $\pm$ 2.34                                                            \\
Sim2Real (with augmentation)    & 1.66 $\pm$ 0.16                                                        & 11.60 $\pm$ 4.65                                                            \\ \hline
Sim+Real (16 real datapoints)             & 1.55 $\pm$ 0.12                                                        & 9.08 $\pm$ 1.65                                                             \\
Sim+Real (32 real datapoints)            & 1.36 $\pm$ 0.05                                                        & 7.95 $\pm$ 1.60                                                             \\
Sim+Real (64 real datapoints)            & 1.24 $\pm$ 0.03                                                        & 8.25 $\pm$ 0.77                                                             \\
Sim+Real (128 real datapoints)           & \textbf{0.52 $\pm$ 0.03}                                               & \textbf{4.14 $\pm$ 0.57}                                                   
\end{tabular}
\caption{Sim2Real experiment results for pose estimation of a cylinder, evaluated in position error of the contact center, and rotation error. We repeat the experiments 5 times, and report the mean error $\pm$ standard deviation. We compare the results of Sim2Sim, Real2Real (with different training samples), and Sim2Real (with/without data augmentation and with different amount of mixed real data).}
\label{tab:sim2real}
\end{table}

The major goal with \simulator{} is to provide a platform to study representations and algorithms for robot learning using touch as a sensor modality, which the researchers can use to test algorithms before experimenting on real sensors/robots.
However, another interesting venue for experimentation with tactile sensors is Sim2Real, where data from the simulator are used to directly train models that are either applied in the real-world or used in conjunction with a small number of real-world data.
Here, we present a proof-of-concept of Sim2Real on a pose-estimation task, where the goal is to estimate the contact center and angle of a pen w.r.t. the sensor being touched.

\textbf{Setting}
We generated in \simulator{} $10,000$ simulated tactile imprints with corresponding poses, and collected $200$ real tactile imprints with manually annotated poses. 
Following, we trained and validated the same convolutional neural networks on several combinations of datasets: 1) Sim2Sim trained and evaluated on simulated data. 2) Real2Real trained and evaluated on real data 3) Sim2Real trained on simulated data and evaluated on real data. 4) Sim+Real trained on simulated data mixed with a small amount of real data and evaluated on real data.

\begin{figure}[t]   
    \centering
    \includegraphics[width=\linewidth]{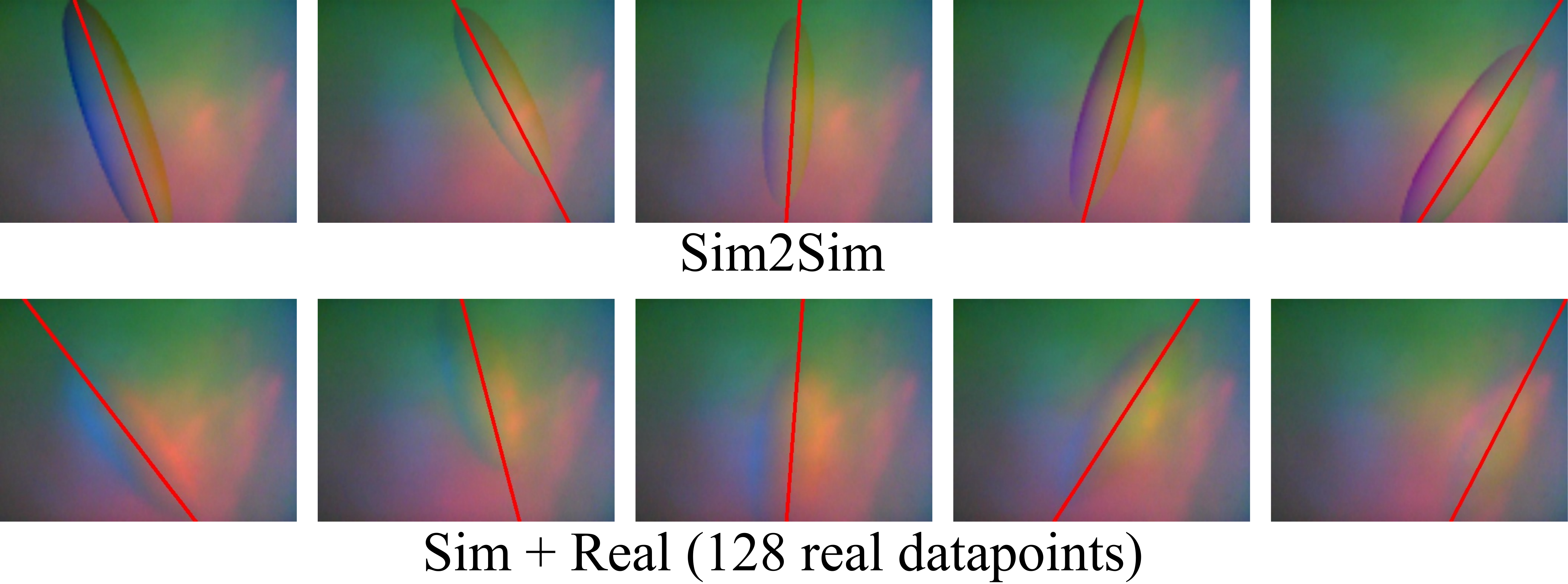}
    \caption{\revision{Examples of tactile reading from simulated (\textit{Top row}) and real data (\textit{Bottom row}), and the corresponding estimated pose (red line).}}
    \label{fig:pose_estimation}
\end{figure}

\textbf{Results}
We show quantitative and qualitative results in \tab{tab:sim2real} and \fig{fig:pose_estimation} respectively. From the results in \tab{tab:sim2real}, we can observe the sim2real gap (Sim2Real without augmentation). 
To bridge the gap, we add data augmentation. We find color jittering very helpful, where we randomly change the brightness and contrast for each RGB channel. It makes the model more robust to a variety of illumination conditions (Sim2Real with augmentation vs. Sim2Real without augmentation). 
We also compare the results between Sim2Real and Real2Real. 
Without any real data, Sim2Real with augmentation can achieve comparable results with Real2Real (64). 
When mixed with real data, Sim2Real consistently outperforms Real2Real with the same amount of real data. These show the potentials for increasing data efficiency by simulated data.

%

%
	

\section{Conclusion}
\label{sec:conclusion}

	In this paper, we introduce \simulator{}, a simulator for vision-based tactile sensors.
\simulator{} is designed to provide an easy-to-use, fast and flexible simulator capable of generating realistic high-resolution readings. 
We demonstrate and validate \simulator{} on a perceptual task for learning grasp stability, and a control task for marble manipulation. 
Moreover, we provide a proof-of-concept that \simulator{} can be successfully used for Sim2Real applications. 
To foster the tactile sensing community and to enable robotics and machine learning researchers to make use of touch in simulation, we open-source \simulator{} at \website{}.

Future work will focus on improving the modeling of the effects of forces through the deformation of the elastomers, and ultimately generating more realistic readings. Besides, different experiment variations can be evaluated, such as the use of different deformation models, the effects of image filtering, the comparison of simulated tactile sensors with real-world sensors on varying geometries, and learning and transferring grasp policies on different objects.

\vspace{-5pt}
\section*{Acknowledgment}

We thank Edward Smith and Huazhe Xu for feedback on early versions of the simulator; Frederik Ebert, Stephen Tian and Akhil Padmanabha for providing the models of the OmniTact; The speakers, participants and organizers of the \textit{ICRA 2020 ViTac Workshop} and the \textit{RSS 2020 Workshop Visuotactile Sensors for Robust Manipulation} for insightful discussions.
\vspace{-12pt}



\bibliographystyle{IEEEtran}
\bibliography{paper}


\end{document}